\documentclass[journal]{IEEEtran}

\usepackage{times}
\usepackage{epsfig}
\usepackage{graphicx}
\usepackage{subcaption}
\usepackage{amsmath}
\usepackage{amssymb}
\usepackage{color}
\usepackage{multirow}
\usepackage{bm}
\usepackage{dsfont}
\usepackage{algorithm}
\usepackage{algorithmic}
\usepackage{array}
\usepackage{textcomp}
\usepackage[normalem]{ulem}
\usepackage{etoolbox}

\graphicspath{{Figures/}}

\usepackage[breaklinks=true,bookmarks=false]{hyperref}

\newcommand{\qed}{\nobreak \ifvmode \relax \else
      \ifdim\lastskip<1.5em \hskip-\lastskip
      \hskip1.5em plus0em minus0.5em \fi \nobreak
      \vrule height0.75em width0.5em depth0.25em\fi}

\usepackage{xspace}

\newcommand{\MyMapTemplatePrefixc}[4]{\expandafter#1\csname#3#4\endcsname{#2{#4}}} 
\forcsvlist{\MyMapTemplatePrefixc {\def} {\mathcal}{c}} {A,B,C,D,E,F,G,H,I,J,K,L,M,N,O,P,Q,R,S,T,U,V,W,X,Y,Z}  

\newcommand{\MyMapTemplatePrefixtb}[5]{\expandafter#1\csname#4#5\endcsname{#2{#3{#5}}}} 
\forcsvlist{\MyMapTemplatePrefixtb {\def} {\tilde}{\mathbf}{t}} {a,b,c,d,e,f,g,h,i,j,k,l,m,n,o,p,q,r,s,t,u,v,w,x,y,z}  

\newcommand{\MyMapTemplateNoPrefix}[3]{\expandafter#1\csname#3\endcsname{#2{#3}}}
\forcsvlist{\MyMapTemplateNoPrefix {\def} {\mathbf}} {0, 1, a, b, c, d, e, f, g, h, i, j, k, l, m, n, o, p, q, r, u, v, w, x, y, z} 
\forcsvlist{\MyMapTemplateNoPrefix {\def} {\mathbf}} {A,B,C,D,E,F,G,H,I,J,K,L,M,N,O,P,Q,R,S,T,U,V,W,X,Y,Z}  

\def\ie{\emph{i.e.}}
\def\eg{\emph{e.g.}}

\def\etal{\emph{et al.}\@\xspace}

\def\ie{\emph{i.e.}\@\xspace}
\def\eg{\emph{e.g.}\@\xspace}
\def\resp{\emph{resp.}\@\xspace}
\def\wrt{\emph{w.r.t.}\@\xspace}

\begin{document}

\title{Learning from Noisy Web Data with Category-level Supervision}

\author{Li Niu, Qingtao Tang, Ashok Veeraraghavan, and Ashu Sabharwal\
\thanks{L. Niu is with Electric and Computer Engineering (ECE) department in Rice University, Houston, TX 77005, USA (e-mail:ln7@rice.edu).}
\thanks{Q. Tang is with Department of Computer Science and Technology, Tsinghua University, Beijing, 100084, China (e-mail:tqt15@mails.tsinghua.edu.cn).}
\thanks{Ashok Veeraraghavan is with Electric and Computer Engineering (ECE) department in Rice University, Houston, TX 77005, USA (e-mail:vashok@rice.edu).}
\thanks{Ashu Sabharwal is with Electric and Computer Engineering (ECE) department in Rice University, Houston, TX 77005, USA (e-mail:ashu@rice.edu).}
}

\maketitle

\begin{abstract}
As tons of photos are being uploaded to public websites (\eg, Flickr, Bing, and Google) every day, learning from web data has become an increasingly popular research direction because of freely available web resources, which is also referred to as webly supervised learning. Nevertheless, the performance gap between webly supervised learning and traditional supervised learning is still very large, owning to the label noise of web data. To be exact, the labels of images crawled from public websites are very noisy and often inaccurate. Some existing works tend to facilitate learning from web data with the aid of extra information, such as augmenting or purifying web data by virtue of instance-level supervision, which is usually in demand of heavy manual annotation. Instead, we propose to tackle the label noise by leveraging more accessible category-level supervision. In particular, we build our method upon variational autoencoder (VAE), in which the classification network is attached on the hidden layer of VAE in a way that the classification network and VAE can jointly leverage the category-level hybrid semantic information.  The effectiveness of our proposed method is clearly demonstrated by extensive experiments on three benchmark datasets.
\end{abstract}

\section{Introduction} \label{sec:intro}
In recent years, we witness the huge success of image classification, which is largely fueled by large-scale publicly available image datasets. Whereas, labeling large-scale dataset is labor-intensive and time-consuming. Thus, it is reasonable that webly supervised learning, \ie, learning from web images, has become increasingly popular due to the abundant freely available web data uploaded to public websites. However, webly supervised learning is encumbered by one critical issue, that is, the labels of web data crawled from public websites are very noisy and often unreliable. Learning classifiers on the noisy web data may lead to significant performance decay on the test set. Lots of research works have been done to cope with the label noise  with various technical approaches~\cite{sukhbaatar2014training,chen2015webly,
bergamo2010exploiting,xiao2015learning,zhuang2016attend,reed2014training}, but webly supervised learning is still struggling to compete with conventional supervised learning. Recently, more and more works begin to consider facilitating webly supervised learning with the aid of extra information, such as purifying or augmenting the web data by selecting informative web data to label~\cite{krause2016unreasonable} or leveraging strong supervision (\eg, clean images, part landmarks, or bounding boxes) from well-labeled dataset~\cite{xu2016webly,xu2016few}. Generally speaking, these approaches aim to clean the noisy labels in the web data or transfer the knowledge from strong supervision to web data. However, they require heavy manual annotation on the instance level, which is usually difficult to obtain. 

In contrast with instance-level supervision, category-level supervision is relatively much easier to acquire in the real-world applications. One of the most well-known category-level supervision is attribute, which is manually designed semantic representation for each category~\cite{farhadi2009describing,lampert2014attribute}. Attribute is usually depicted in the form of real-valued vector with each entry representing specific information such as shape (\eg, cylindrical), material (\eg, cloth), and color (\eg, white).
When attribute is unavailable, an alternative is the word vector corresponding to each category name. Specifically, a linguistic model is trained based on free online corpus (\eg, Wikipedia)~\cite{akata2015evaluation, frome2013devise, socher2013zero, yu2013designing}, and a real-valued vector can be obtained for each word, which is dubbed word vector. The word vector corresponding to each category name has been extensively employed as category-level semantic information in existing zero-shot learning works~\cite{akata2015evaluation, frome2013devise, socher2013zero, yu2013designing}.
Besides attribute and word vector, category-level semantic information can also be obtained by summarizing the visual features of web images despite their inaccurate labels. Specifically, we propose a novel category-level semantic representation named visual encoding, by averaging the visual features within each category as visual encoding, followed by enhancement strategies which enable the visual encoding to be more robust and discriminative. More details can be found in Section~\ref{sec:csi}.
Considering the availability of category-level semantic information, we  explore learning from web data with category-level supervision, which is located in the middle ground between instance-level supervision and no extra supervision at all.

\begin{figure*}[t]
	\centering
    \includegraphics[width=0.95\textwidth]{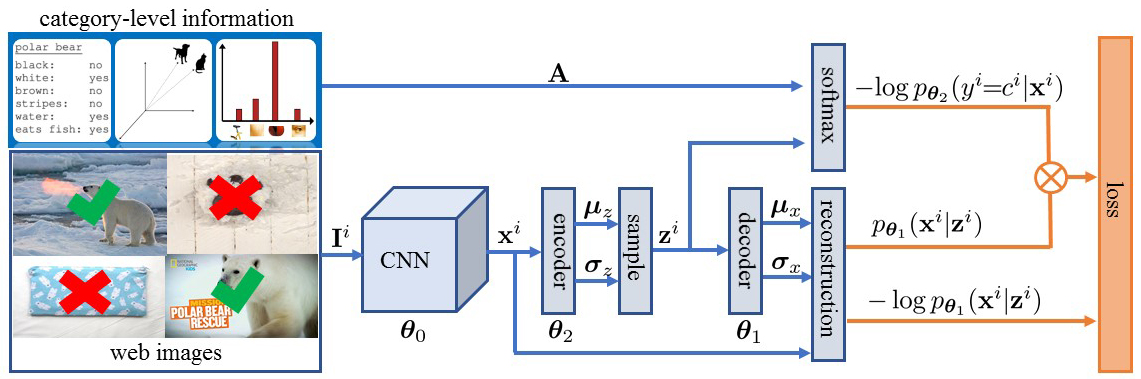}              
    \caption{Flowchart of learning from noisy web data with category-level semantic information. The top flow is classification network and the bottom flow is variational autoencoder. Two flows share the common model parameters $\bm{\theta}_0$ and $\bm{\theta}_2$.}
    \label{fig:flowchart}
\end{figure*}

With the aim to cope with label noise with category-level supervision, we base our design on autoencoder-like network structure, considering that autoencoder has been used for outlier detection based on reconstruction error~\cite{sakurada2014anomaly,xia2015learning} and its hidden layer can be regulated by prior information, which meets the requirement of our task. In particular, our probabilistic method is built upon a probabilistic variant of autoencoder, \ie, variational autoencoder (VAE)~\cite{kingma2013auto,rezende2014stochastic}, considering that the reconstruction probability density provided by VAE can be readily integrated into our probabilistic method.  Nonetheless, outlier detection is a non-trivial problem for VAE with multi-category noisy training data. To solve this problem, we propose our method named Webly Supervised learning with Category-level Information (WSCI), which is illustrated in Figure~\ref{fig:flowchart}. From Figure~\ref{fig:flowchart}, it can be seen that our network consists of a classification network in the top flow and variational autoencder (VAE) in the bottom flow, which share common modules (\ie, CNN and encoder) and jointly utilize category-level information. The classification network is attached on the hidden layer of VAE and incorporates category-level information to generate prediction scores. In this sense, the latent variables in the hidden layer of VAE are endowed with semantic meanings. Hence, we refer to such VAE as semantic VAE and the latent variables of VAE as semantic embeddings. When training our WSCI model, the classification network and VAE influence each other in the following way. On one hand, VAE identifies outliers to assist in learning a more robust classification network by assigning higher weights on the losses of identified non-outliers. On the other hand, the classification network injects relatively accurate discriminative information into the hidden layer of VAE, contributing to a stronger semantic VAE for outlier detection.

Our major contributions can be summarized as follows: 1) as far as we are concerned, this is the first VAE based work to cope with the label noise when learning from web data by utilizing hybrid semantic information as category-level supervision; 2) we propose a deep probabilistic WSCI method, in which classification model and VAE can jointly leverage category-level information; 3) extensive experiments show the superiority of our proposed method for learning from web data.

\section{Related Work} \label{sec:related}
\noindent\textbf{Webly Supervised Image Classification: }
In recent years, a great many research works~\cite{chen2013neil,bergamo2010exploiting,niu2015visual,
li2014exploiting, niu2016exploiting, niu2017action,niu2016visual} were proposed to handle the label noise when learning from web data, by utilizing weakly supervised learning techniques. More recently, several deep learning approaches were developed for webly supervised learning~\cite{xiao2015learning,
sukhbaatar2014training,chen2015webly,zhuang2016attend,reed2014training,divvala2014learning}
by using label flip layer, bootstrapping, query expansion, attention model, or multi-instance learning techniques. The above works are capable of mitigating the label noise to some extent when learning from web data. Nevertheless, the performance gap between webly supervised learning and traditional supervised learning remains very large.

With the goal to close this performance gap, some recent research works relied on auxiliary information such as selective labeling~\cite{krause2016unreasonable} or extraneous strong supervision~\cite{xu2016webly,xu2016few} (\eg, clean images, part landmarks, or bounding boxes), which requires manual labeling on the instance level. Roughly speaking, these works aimed to clean the noisy labels of web data with human intervention or transfer the knowledge from strong supervision to web data. However, instance-level annotation is often difficult to obtain in the real-world application. Instead, we intend to use more accessible category-level information for enhancing the performance of webly supervised learning. Some existing works~\cite{chen2015webly,xiao2015learning} are equipped with the potential to utilize the similarities among different categories based on category-level information, whereas they are not able to take full advantage of category-level information.

\noindent\textbf{Variational Autoencoder: }
Variational autoencoder (VAE) \cite{kingma2013auto,rezende2014stochastic} is a probabilistic generative model (the technical details of VAE is left to Section~\ref{sec:vae}). 
Recently, one popular research direction \wrt VAE is to regulate the hidden layer of VAE more heavily. To name a few, conditional VAE (CVAE)~\cite{sohn2015learning,walker2016uncertain,yan2016attribute2image} simply feeds extra prior information to the input or hidden layer of VAE, adversarial autoencoder (AAE)~\cite{makhzani2015adversarial} attaches a generative adversarial network (GAN) on top of the hidden layer of VAE, and semi-supervised VAE~\cite{kingma2014semi} utilizes both labeled and unlabeled data in the training stage. 
In terms of regulating the hidden layer of VAE, our WSCI method is also within this scope. However, our motivation of regulating the hidden layer of VAE is to tackle the label noise in training data, which is different from the motivations of all the works enumerated above~\cite{sohn2015learning,walker2016uncertain,yan2016attribute2image,
makhzani2015adversarial,kingma2014semi}.

Although a few works ~\cite{an2015variational, suh2016echo} have exploited VAE for outlier detection, but they lacked the discussion on handling the label noise in the multi-category training data. Moreover, they failed to directly yield a robust classifier. In the contrast, our WSCI method can tackle the label noise in the multi-category training data and simultaneously produce a roust classifier.

\setlength{\textfloatsep}{0pt}

\section{Background}\label{sec:vae} 
In the remainder of this paper, for the convenience of representation, we denote a matrix/vector by using a uppercase/lowercase letter in boldface (\eg, $\A$ denotes a matrix and $\a$ denotes a vector).
We use $\I$ to denote identity matrix and $\1$ (\resp, $\0$) to denote all-one (\resp, all-zero) vector/matrix when its dimension is obvious. Besides,
$\A^T$ (\resp, $\A^{-1}$) is used to denote the transpose (\resp, inverse) of $\A$. Moreover, we use $\A\circ\B$ (\resp, $\langle\A,\B\rangle$) to denote the element-wise (\resp, inner) product between $\A$ and $\B$.

Now we introduce the background knowledge of variational autoencoder (VAE), which our probabilistic method are built upon. 
Suppose that instance $\x$ can be generated from latent variable $\z$, then the marginal likelihood of $\x$ can be represented as
$p_{\bm{\theta}_1}(\x)=\int p_{\bm{\theta}_1}(\x|\z)p_{\bm{\theta}_1}(\z)d\z$ with generative parameters $\bm{\theta}_1$, in which $p_{\bm{\theta}_1}(\x|\z)$ is the likelihood of $\x$ given $\z$ and $p_{\bm{\theta}_1}(\z)$ is the prior over latent variable $\z$.

Nevertheless, $\int p_{\bm{\theta}_1}(\x|\z)p_{\bm{\theta}_1}(\z)d\z$ is intractable over all configurations of latent variables. In order to address this issue, variational autoencoder (VAE)~\cite{kingma2013auto,rezende2014stochastic} introduces approximate posterior $q_{\bm{\theta}_2}(\z|\x)$ with variational parameters $\bm{\theta}_2$.  Then, instead of maximizing the marginal likelihood $p_{\bm{\theta}_1}(\x)$, VAE proposes to maximize the lowerbound of marginal likelihood $p_{\bm{\theta}_1}(\x)$, \ie, Evidence Lower BOund (ELBO), which is equivalent to minimizing the following objective function (please refer to \cite{kingma2013auto} for the detailed derivation):
\begin{eqnarray}
\textnormal{KL}[q_{\bm{\theta}_2}(\z|\x)||p_{\bm{\theta}_1}(\z)]-\mathbb{E}_{q_{\bm{\theta}_2}(\z|\x)} [\log p_{\bm{\theta}_1}(\x|\z)], \label{eqn:ELBO}
\end{eqnarray}
\noindent in which the first regularizer enforces the learnt approximate posterior $q_{\bm{\theta}_2}(\z|\x)$ to be close to the given prior $p_{\bm{\theta}_1}(\z)$ by minimizing the KL divergence between  $q_{\bm{\theta}_2}(\z|\x)$ and $p_{\bm{\theta}_1}(\z)$, and the second regularizer suppresses reconstruction error by maximizing the expectation of $\log p_{\bm{\theta}_1}(\x|\z)$ \wrt $q_{\bm{\theta}_2}(\z|\x)$. In other words, the objective function in (\ref{eqn:ELBO}) tends to learn an approximate posterior close to the given prior and encourage the truthfulness of reconstruction at the same time. Regarding the explicit forms of $p_{\bm{\theta}_1}(\z)$,  $p_{\bm{\theta}_1}(\x|\z)$, and $q_{\bm{\theta}_2}(\z|\x)$, it is commonly assumed that $p_{\bm{\theta}_1}(\z)=\mathcal{N}(\z; \0, \I)$ for simplicity. Furthermore, following~\cite{kingma2013auto}, we make the assumption that $p_{\bm{\theta}_1}(\x|\z)$ (\resp, $q_{\bm{\theta}_2}(\z|\x)$) is a multivariate Gaussian with diagonal covariance, \ie, $p_{\bm{\theta}_1}(\x|\z) = \mathcal{N}(\x; \bm{\mu}_x, \textnormal{diag}(\bm{\sigma}_x^2))$ (\resp, $q_{\bm{\theta}_2}(\z|\x) = \mathcal{N}(\z; \bm{\mu}_z, \textnormal{diag}(\bm{\sigma}_z^2))$), in which $p_{\bm{\theta}_1}(\x|\z)$ (\resp, $q_{\bm{\theta}_2}(\z|\x)$) is specified by a probabilistic decoder (\resp, encoder) network with model parameters $\bm{\theta}_1$ (\resp, $\bm{\theta}_2$).

The forward procedure of VAE is made up of three steps: 1) generate the variational parameters $\bm{\mu}_z$ and $\bm{\sigma}_z$ of $q_{\bm{\theta}_2}(\z|\x)$ (approximate posterior) based on $\x$ using probabilistic encoder; 2) sample latent variables $\z$ based on the generated $\bm{\mu}_z$ and $\bm{\sigma}_z$; 3) generate the generative parameters $\bm{\mu}_x$ and $\bm{\sigma}_x$ of $p_{\bm{\theta}_1}(\x|\z)$ (likelihood of $\x$ given $\z$) based on the sampled $\z$ using probabilistic decoder. The above process is illustrated in the bottom flow in Figure~\ref{fig:flowchart}. In the second step of forward process, reparameterization trick is usually employed for ease of optimization~\cite{kingma2013auto}. To be exact, instead of sampling directly from approximate posterior $q_{\bm{\theta}_2}(\z|\x)=\mathcal{N}(\z; \bm{\mu}_z,\textnormal{diag}(\bm{\sigma}_z^2))$, deterministic mapping $\z=g_{\bm{\theta}_2}(\x,\bm{\epsilon})=\bm{\mu}_z+\bm{\sigma}_z\circ\bm{\epsilon}$ with $\bm{\epsilon}\sim\mathcal{N}(\0,\I)$ can be used to obtain $\z$, because it has been proved that
$\mathbb{E}_{\z\sim\mathcal{N}(\bm{\mu}_z,\textnormal{diag}(\bm{\sigma}_z^2))} f(\z)=\mathbb{E}_{\bm{\epsilon}\sim\mathcal{N}(\0,\I)} f(\bm{\mu}_z+\bm{\sigma}_z\circ\bm{\epsilon})\approx \frac{1}{L}\sum_{l=1}^L f(\bm{\mu}_z+\bm{\sigma}_z\circ\bm{\epsilon}^l)$ with $L$ being the number of samples per training instance~\cite{kingma2013auto}. In practice, as long as the batch size is large enough, each instance can be sampled only once (\ie, $L=1$) in each training epoch~\cite{kingma2013auto}.


\section{Webly Supervised Learning with Category-level Information} \label{sec:WSCI}
In this section, we propose our method named Webly Supervised learning with Category-level Information (WSCI) based on variational autoencoder (VAE),  in which the classification network and VAE can jointly utilize category-level information to cope with the label noise. 

\subsection{Semantic VAE for Outlier Detection} \label{sec:vae_outlier}

For better explanation, we rewrite the objective function of VAE in (\ref{eqn:ELBO}) as follows,
\begin{eqnarray} \label{eqn:VAE_optimization_ori}
\textnormal{KL}[q_{\bm{\theta}_2}(\z|\x)||p_{\bm{\theta}_1}(\z)]-\mathbb{E}_{q_{\bm{\theta}_2}(\z|\x)} [\log p_{\bm{\theta}_1}(\x|\z)].
\end{eqnarray}
\noindent Similar to autoencoder used for outlier detection, VAE has also been used for detecting outliers based on the reconstruction probability density $p_{\bm{\theta}_1}(\x|\z)$~\cite{an2015variational,suh2016echo}. Specifically, an instance $\x^i$ is supposed to be an outlier when $\mathbb{E}_{q_{\bm{\theta}_2}(\z|\x^i)}p_{\bm{\theta}_1}(\x^i|\z)$ is below certain threshold. 
Ideally, to learn a normal profile for non-outliers, VAE is expected to be trained on clean training data, which is not applicable in our scenario as our web training data are quite noisy. One possible solution is that we can learn one VAE for each category, because the distribution within each category is relatively coherent and training on coherent data can mitigate the negative effect of outliers to some degree, referring to the explanation in~\cite{xia2015learning}. However, this solution is rather cumbersome especially when there are a large number of categories. As an alternative approach, we intend to inject category-level semantic information into the hidden layer and learn a semantic VAE. Although a similar concept, \ie, semantic autoencoder, has been introduced in~\cite{kodirov2017semantic}, the work in~\cite{kodirov2017semantic} focuses on the projection domain shift problem between seen categories and unseen categories in zero-shot learning while our work focuses on the label noise of web training data.  The motivation and details of our semantic VAE will be fully described later in this section.

Given a noisy training set $\mathcal{I}=\{\I^i|_{i=1}^n\}$ with $\I^i$ being the $i$-th image and $n$ being the total number of training images, we denote the visual feature of $\I^i$ (\ie, output of CNN in Figure~\ref{fig:flowchart}) as $\x^i$ and its associated label (\resp, prediction variable from the classification network in Figure~\ref{fig:flowchart}) as $c^i$ (\resp, $y^i$). Recall that per training epoch, only one latent variable is sampled for each training instance using reparameterization trick (see Section~\ref{sec:vae}), so we use  $\z^i=g_{\bm{\theta}_2}(\x^i,\bm{\epsilon}^1)$  to denote the deterministic latent variable of $\x^i$. Then, the objective function in (\ref{eqn:VAE_optimization_ori}) \wrt individual $\x^i$ can be written as
\begin{eqnarray} \label{eqn:VAE_optimization}
\textnormal{KL}[q_{\bm{\theta}_2}(\z|\x^i)||p_{\bm{\theta}_1}(\z)]- \log p_{\bm{\theta}_1}(\x^i|\z^i).
\end{eqnarray}

Next, we will introduce how to inject category-level semantic information into the hidden layer of VAE to form semantic VAE, after which $\z^i$ is expected to represent the semantic embedding of $\x^i$. By denoting $\mathcal{X}^{\tilde{c}}=\{\x^i|c^i=\tilde{c}\}$ and $\mathcal{Z}^{\tilde{c}}=\{\z^i|\x^i\in \mathcal{X}^{\tilde{c}}\}$, each $\mathcal{Z}^c$ corresponding to each category should be densely distributed. In lieu of enforcing the distribution of $\mathcal{Z}^c$ to be close to certain hypothetical distribution like previous VAE based works~\cite{kingma2013auto,rezende2014stochastic,makhzani2015adversarial}, we tend to regulate the hidden layer indirectly inspired by recent ZSL works~\cite{akata2015evaluation,romera2015embarrassingly,
niu2017zero}, which compute the prediction scores based on instance-level semantic embeddings and category-level semantic presentations (\eg, attribute vector).
In particular, in ZSL works~\cite{akata2015evaluation,romera2015embarrassingly}, the prediction score of $\x^i$ is calculated by using ${\x^i}^T\V\A$, in which $\V\in\mathcal{R}^{d\times m}$ is mapping matrix with $d$ (\resp, $m$) being the dimension of visual feature (\resp, attribute vector), and $\A\in\mathcal{R}^{m\times C}$ is category-attribute matrix with the $c$-th column $\a^c$ being the attribute vector of the $c$-th category for a total number of $C$ categories.
In analogy to ${\x^i}^T\V\A$, we expect ${\z^i}^T\A$ ($\z^i$ is similar to the semantic embedding ${\x^i}^T\V$) to be consistent with the prediction score of $\x^i$, where $\A$ used in our experiments is hybrid category-level semantic representations and will be detailed in Section \ref{sec:csi}. 
To achieve this goal, we attach a classifier on the hidden layer of VAE (see Figure~\ref{fig:flowchart}) and calculate the predicted category probability of $\z^i$ by $p(y^i=c^i|\z^i) = \frac{\exp({\z^i}^T\a^{c^i})}{\sum_{\tilde{c}=1}^C\exp({\z^i}^T\a^{\tilde{c}})}$. Then, we replace the KL divergence regularizer $\textnormal{KL}[q_{\bm{\theta}_2}(\z|\x^i)||p_{\bm{\theta}_1}(\z)]$ in (\ref{eqn:VAE_optimization}) with softmax loss $-\log p(y^i=c^i|\z^i)$, leading to the following loss function of our semantic VAE:
\begin{eqnarray}\label{eqn:loss_lb_noisy}
l_{\bm{\theta}}(\x^i,c^i)=-\log p(y^i=c^i|\z^i)-\log p_{\bm{\theta}_1}(\x^i|\z^i),
\end{eqnarray}
\noindent in which $\bm{\theta}=\{\bm{\theta}_0, \bm{\theta}_1,\bm{\theta}_2\}$ including the CNN parameters $\bm{\theta}_0$, generative parameters $\bm{\theta}_1$, and variational parameters $\bm{\theta}_2$. 
Note that the softmax loss in (\ref{eqn:loss_lb_noisy}) is an empirical approach to regulate the latent variables of VAE, sharing a similar spirit with the KL divergence regularizer in VAE, which is also used to regulate the latent variables. However, unlike the KL divergence regularizer in VAE, which imposes non-discriminative prior on the latent variables, the softmax loss in our semantic VAE distinguish the latent variables corresponding to the instances from different categories, making $p_{\bm{\theta}_1}(\x^i|\z^i)$ more reliable for indicating non-outliers than that in VAE. The reasons can be explained as below. 

First, based on the analysis in~\cite{xia2015learning}, when using gradient descent to minimize the reconstruction error, the learnt model is more capable of detecting outliers if the gradients of non-outliers are more consistent.
In our problem, considering each $\mathcal{Z}^c$, due to the first regularizer in (\ref{eqn:loss_lb_noisy}), the distribution of $\mathcal{Z}^c$ in semantic VAE should be more concentrated than that in VAE. Hence,
when minimizing the reconstruction error $\sum_{\z^i\in\mathcal{Z}^c}-\log p_{\bm{\theta}_1}(\x^i|\z^i)$ via gradient descent, the gradients of non-outliers in semantic VAE should be more consistent than those in VAE.
Combined with the analysis in~\cite{xia2015learning}, we can know that compared with VAE, semantic VAE is more adept at detecting outliers and the corresponding $p_{\bm{\theta}_1}(\x^i|\z^i)$ is a more reliable indicator of $\x^i$ being a non-outlier.

In Section~\ref{sec:vae}, we mention that (\ref{eqn:ELBO}) is the upperbound of $− \log p_{\bm{\theta}}(\mathbf{x})$ in VAE. In our semantic VAE, although there is no such theoretical guarantee, the effectiveness of our method is not encumbered in practice, which will be shown in our experimental section (see Section~\ref{sec:exp}). In fact, the effectiveness of replacing the KL divergence in an empirical way without theoretical guarantee has also been proved in some previous research works like AAE~\cite{makhzani2015adversarial}, in which the latent variables of VAE are regulated by using an adversarial training process. In the next section, our emphasis will be transferred from semantic VAE to classifier. To be exact, we will explore how to learn a robust classifier with the aid of semantic VAE and how the robust classifier can affect semantic VAE in return. In the rest part of this paper, to avoid ambiguity, ``VAE" means semantic VAE by default and original VAE without semantic information is differentiated with the name ``plain VAE" instead.

\subsection{Learn Robust Classifier with Label Noise} \label{sec:robust_classifier}
In this section, we target at learning a robust classifier with the assistance of the reconstruction probability density from VAE. Recall that we have already attached a classifier on the hidden layer of VAE to regulate the latent variables in Section~\ref{sec:vae_outlier}. As opposed to adding a totally separate classification network for prediction, we choose to tweak the attached classifier to deal with the label noise. One benefit of doing so is the reduced number of parameters to be learnt through sharing model parameters and another byproduct is a more effective VAE, which will be elaborated later in this section.

In order to tackle the label noise, we introduce a new variable $\tilde{y}^i$ to indicate the noisy label variable of $\x^i$, which is different from the predicted label variable $y^i$ of $\x^i$. Moreover, we further introduce a binary hidden variable $h^i$ to indicate whether the instance $\x^i$ is an outlier or not, that is, $h^i=1$ if $\x^i$ is a non-outlier and $h^i=0$ otherwise. In a strict sense, label noise of web data can be divided into two types: outlier noise (image belongs to none of the training categories) and  label flip noise (image belongs to one of the other training categories)~\cite{sukhbaatar2014training}. However, according to the claim in \cite{xu2016webly} as well as our own experimental observation, outlier noise is far more dominant than label flip noise. Thus, we simply treat all the label noise as outlier noise without considering label flip noise separately in this work.
 
After defining the noisy label variable $\tilde{y}^i$, let us consider the conditional probability of noisy label, \ie, $p_{\bm{\theta}}(\tilde{y}^i|\x^i,h^i)$. Since we treat all the label noise as outlier noise as mentioned above, when $h^i=0$, $\x^i$ is not from any training category and likely to be randomly assigned with arbitrary category label, so $p_{\bm{\theta}}(\tilde{y}^i|\x^i,h^i=0)=\frac{1}{C}$. When $h^i=1$, the associated label of $\x^i$ is correct and we expect $y^i$ predicted by our model to be compatible with $\tilde{y}^i$. In summary, $p_{\bm{\theta}}(\tilde{y}^i|\x^i,h^i)$ can be characterized as
\begin{eqnarray} \label{eqn:y_assump}
p_{\bm{\theta}}(\tilde{y}^i|\x^i,h^i) = 
\begin{cases}
\frac{1}{C},\quad \textnormal{if}\,\, h^i=0,\\
p_{\bm{\theta}}(y^i|\x^i),\quad \textnormal{if}\,\, h^i=1.
\end{cases}
\end{eqnarray}

To this end, we tend to tweak the loss function in (\ref{eqn:loss_lb_noisy}) by taking the label noise into consideration. Particularly, we replace $\log p(y^i=c^i|\z^i)$ in (\ref{eqn:loss_lb_noisy}) with  $\log p(\tilde{y}^i=c^i|\z^i)$, resulting in the following new loss function:
\begin{eqnarray}\label{eqn:loss_lb_ex}
l'_{\bm{\theta}}(\x^i,c^i)=-\log p(\tilde{y}^i=c^i|\z^i)-\log p_{\bm{\theta}_1}(\x^i|\z^i).
\end{eqnarray}

\noindent With deterministic latent variable $\z^i$ given $\x^i$, $p(\tilde{y}^i|\z^i)$ can be approximated by $p_{\bm{\theta}_2}(\tilde{y}^i|\x^i)$, so the loss function in (\ref{eqn:loss_lb_ex}) can be derived as
\begin{eqnarray}
l'_{\bm{\theta}}(\x^i,c^i)\!\!\!\!\!\!\!\!\!\!&&=-\log p_{\bm{\theta}_2}(\tilde{y}^i\!=\!c^i|\x^i)-\log p_{\bm{\theta}_1}(\x^i|\z^i)\nonumber\\
&&\!\!\!\!\!\!\!\!\!\!\!\!\!\!\!\!\!\!\!\!=-\log \sum_{h^i}p_{\bm{\theta}_2}(\tilde{y}_i\!=\!c^i,h^i|\x^i)-\log p_{\bm{\theta}_1}(\x^i|\z^i)\nonumber\\
&&\!\!\!\!\!\!\!\!\!\!\!\!\!\!\!\!\!\!\!\!=-\log\sum_{h^i}p_{\bm{\theta}_2}(\tilde{y}^i\!=\!c^i|\x^i,h^i)p(h^i|\x^i)\!-\!\log p_{\bm{\theta}_1}(\x^i|\z^i)\nonumber\\
&&\!\!\!\!\!\!\!\!\!\!\!\!\!\!\!\!\!\!\!\!\leq -\sum_{h^i}p(h^i|\x^i)\log p_{\bm{\theta}_2}(\tilde{y}^i\!=\!c^i|\x^i,h^i)\nonumber\\
&&\!\!\!\!\!\!\!\!\!\!\!\!\!\!-\log p_{\bm{\theta}_1}(\x^i|\z^i)\label{eqn:jensen_inequal}\\
&&\!\!\!\!\!\!\!\!\!\!\!\!\!\!\!\!\!\!\!\!=-p(h^i\!=\!1|\x^i)\log p_{\bm{\theta}_2}(y^i\!=\!c^i|\x^i)\!-\!\log p_{\bm{\theta}_1}(\x^i|\z^i)\nonumber\\
&&\!\!\!\!\!\!\!\!\!\!\!\!\!\!-(1-p(h^i\!=\!1|\x^i))\log\frac{1}{C}\label{eqn:h_def}\\
&&\!\!\!\!\!\!\!\!\!\!\!\!\!\!\!\!\!\!\!\!=-p(\x^i|h^i\!=\!1)\frac{p(h^i\!=\!1)}{p(\x^i)}\left(\log p_{\bm{\theta}_2}(y^i\!=\!c^i|\x^i)\!+\!\log C\right)\nonumber\\
&&\!\!\!\!\!\!\!\!\!\!\!\!\!\!-\log p_{\bm{\theta}_1}(\x^i|\z^i)+\textnormal{const},\label{eqn:loss_lb_final}
\end{eqnarray}

\noindent where (\ref{eqn:jensen_inequal}) is obtained based on Jensen's inequality and (\ref{eqn:h_def}) is obtained by substituting the definition in (\ref{eqn:y_assump}). Recall that $p_{\bm{\theta}_1}(\x^i|\z^i)$ is a reliable indicator of $\x^i$ being a non-outlier as discussed in Section~\ref{sec:vae_outlier}, which aligns with the meaning of $p(\x^i|h^i=1)$. Hence, we approximate  $p(\x^i|h^i=1)$ in (\ref{eqn:loss_lb_final}) with $p_{\bm{\theta}_1}(\x^i|\z^i)$. After omitting the constant terms, we rewrite (\ref{eqn:loss_lb_final}) as
\begin{eqnarray} \label{eqn:near_final_obj}
l'_{\bm{\theta}}(\x^i,c^i) \propto \!\!\!\!\!\!\!\!\!\!\!\!&&-p_{\bm{\theta}_1}(\x^i|\z^i)\left(\log p_{\bm{\theta}_2}(y^i=c^i|\x^i)+\log C\right)\nonumber\\
&&-\lambda_i\log p_{\bm{\theta}_1}(\x^i|\z^i),
\end{eqnarray}
\noindent where $\lambda_i=\frac{p(\x^i)}{p(h^i=1)}$. As $p(\x^i)$ and $p(h^i=1)$ cannot be directly inferred based on our model, we simply unify all $\lambda_i$'s as a common parameter $\lambda$. An interesting observation from the perspective of classification loss is that $-p_{\bm{\theta}_1}(\x^i|\z^i)\log p_{\bm{\theta}_2}(y^i=c^i|\x^i)$ in (\ref{eqn:near_final_obj}) is essentially weighted softmax loss, which tends to assign higher weights on the training instances that are more likely to be non-outliers. 

Recall that at the beginning of this section, we claim that one byproduct of tweaking the attached classifier is a more effective VAE. Now let us switch to the perspective of VAE. The corresponding $\z^i$'s of identified non-outliers (\ie, with larger $p_{\bm{\theta}_1}(\x^i|\z^i)$) are regulated more heavily than those of identified outliers. In this sense, the distribution of each $\mathcal{Z}^c$ is biased towards identified non-outliers from $\mathcal{X}^c$ and hence expected to be closer to the distribution of ground-truth $\bar{\mathcal{Z}}^c$, which is the semantic embedding space of ground-truth non-outliers from $\mathcal{X}^c$. When the distribution of $\mathcal{Z}^c$ moves closer to that of $\bar{\mathcal{Z}}^c$, the corresponding $\z^i$'s of ground-truth non-outliers from $\mathcal{X}^c$ are supposed to be more densely distributed. Thus, their gradients when minimizing the reconstruction error should be more consistent. Similar to the discussion in Section~\ref{sec:vae_outlier}, more consistent gradients of ground-truth non-outliers will lead to better capability of VAE for detection outliers. Intuitively, through injecting relatively accurate discriminative information into the hidden layer, we can achieve a more effective VAE.

Based on the above analyses from the perspectives of both classification loss and semantic VAE, the classification network and semantic VAE influence each other, leading to a more robust classifier and a more effective VAE.
The loss function in (\ref{eqn:near_final_obj}) can be minimized by training an end-to-end deep learning system as shown in Figure~\ref{fig:flowchart}. However, our training process is stunted by two practical issues: 1) The first issue is that for high-dimension multivariate Gaussian distribution, the value of probability density function is generally too small or too large, resulting in unexpected numerical problems. Thus, in lieu of directly calculating $p_{\bm{\theta}_1}(\x^i|\z^i)$, we first calculate $\log p_{\bm{\theta}_1}(\x^i|\z^i)$, then subtract the maximum value from each training batch, and finally compute the exponential to recover $p_{\bm{\theta}_1}(\x^i|\z^i)$. In this way, we can obtain the normalized $p_{\bm{\theta}_1}(\x^i|\z^i)$ in the range of $[0,1]$, which is denoted as $\tilde{p}_{\bm{\theta}_1}(\x^i|\z^i)$. This trick is equal to multiplying the first term in (\ref{eqn:near_final_obj}) with a constant per training batch, similar in spirit to batch normalization~\cite{ioffe2015batch}. Note that the multiplied constant can be approximately absorbed into the trade-off parameter $\lambda$, resulting in a new trade-off parameter $\tilde{\lambda}$; 2) The second issue is that $\log p_{\bm{\theta}_1}(\x^i|\z^i)$'s are subject to such high variance that in each training batch, only one $\tilde{p}_{\bm{\theta}_1}(\x^i|\z^i)$ reaches $1$ while all the others are pushed to $0$. To circumvent this issue, $\bm{\sigma}_x$ is fixed as $\1$ so that $p_{\bm{\theta}_1}(\x|\z)=\mathcal{N}(\x;\bm{\mu}_x,\I)$, which makes $\log p_{\bm{\theta}_1}(\x^i|\z^i)$'s more stable. After addressing the above two issues, we arrive at our final optimization problem:
\begin{eqnarray} \label{eqn:final_obj}
\min_{\bm{\theta}} \sum_{i=1}^n\!\!\!\!\!\!\!\!\!\!\!\! && -\tilde{p}_{\bm{\theta}_1}(\x^i|\z^i) \left(\log p_{\bm{\theta}_2}(y^i\!=\!c^i|\x^i)+\log C\right)\nonumber\\
&&-\tilde{\lambda}\log p_{\bm{\theta}_1}(\x^i|\z^i),
\end{eqnarray}

\noindent in which $n$ is the number of training instances.

In the testing phase, given a test instance $\x^i$, we use $\mathbb{E}_{q_{\bm{\theta}_2}(\z|\x^i)} p(y^i|\z)$ $\approx$ $\frac{1}{L} \sum_{l=1}^L p(y^i|\z^{i,l})$ with $\z^{i,l}=g_{\bm{\theta}_2}(\x^i,\bm{\epsilon}^l)$ (see Section~\ref{sec:vae}) for prediction. In particular, each test instance is passed through the classification network for $5$ times (\ie, $L=5$) and the averaged category probabilities is used as its final category probability. 

\noindent\textbf{Discussion on two noise types: }As mentioned before, 
strictly speaking, label noise of web data can be categorized into outlier noise and label flip noise~\cite{sukhbaatar2014training}. However, in this work, all the label noise are simply treated as outlier noise and the label flip noise is not considered separately, due to the following concerns: 1) In theory, in our VAE based method, the mathematical derivation could be greatly eased by treating all the label noise as outlier noise, whereas it is a far more challenging task to distinguish two types of noise under our VAE based framework; 2) In practice, regardless of the concrete noise type
(outlier noise or label flip noise), all noisy images are supposed to
have lower weights $p_{\bm{\theta}}(x_i| z_i)$ based on the analysis in the paragraph below (\ref{eqn:loss_lb_noisy}). Thus, all noisy images will be suppressed with lower weights during the training procedure of minimizing (\ref{eqn:final_obj}), which contributes to learning a more robust classifier; 3) Some previous works like~\cite{xiao2015learning} consider the label flip noise separately and flip the mistaken labels to correct labels, which can make the most of training instances. Nonetheless, according to our experimental observation, the instances
with label flip noise are very limited, which is the same as the observation in~\cite{xu2016webly}.

\section{Category-level Semantic Representation} \label{sec:csi}
Last but not least, we introduce the category-level semantic representation used in our experiments (matrix $\A$ in Section \ref{sec:WSCI}), which contains three types of information. As mentioned in Section~\ref{sec:intro}, two commonly used types of category-level semantic information are attribute and word vector. However, attribute calls for the manual annotation of human experts, which is often inaccessible, while word vector is not grounded on visual information, which may not be suitable for computer vision applications. To overcome the drawbacks of attribute (\eg, not free) and word vector (\eg, free yet not visually grounded), we propose a third type of category-level semantic information called visual encoding, which is both free and visually grounded.

\noindent\textbf{Attribute: }Attribute representation~\cite{lampert2014attribute,farhadi2009describing} for each category is a
high-level description in the form of real-valued vector, with each entry representing specific information such as shape (\eg, cylindrical), material (\eg, cloth), and color (\eg, white). Such attribute representations are in high demand of expertise from human experts, and thus not easy to acquire. 

\noindent\textbf{Word Vector: }Each word can be represented by a real-valued vector (\eg, Word2Vec~\cite{mikolov2013distributed} and GloVe~\cite{pennington2014glove}) by training a linguistic model on free online corpus (\eg, Wikipedia). Word vector prevails in the research community of natural language processing (NLP), and also turns increasingly popular in the field of computer vision such as zero-shot learning (ZSL) approaches~\cite{akata2015evaluation, frome2013devise, socher2013zero, yu2013designing} for image classification.
However, word vector focuses on linguistic regularities and patterns, and hence may lack visual grounding~\cite{Kottur_2016_CVPR}.

\noindent\textbf{Visual Encoding: }We additionally design a free and visually grounded encoding method, which encodes each category as visual bag-of-word representation by using Gaussian Mixture Model (GMM) as the visual codebook.
In particular, we first generate $200$ region proposals for each training image using selective search~\cite{uijlings2013selective} and extract $2048$-dim visual features for all region proposals using pretrained Inception-V3~\cite{szegedy2016rethinking}. Then, we use $1,000,000$ sampled region proposals to train a $K$-component GMM with model parameters $\{\pi_1,\bm{\mu}_1,\bm{\sigma}_1^2;\ldots;\pi_K,\bm{\mu}_K,\bm{\sigma}_K^2\}$, in which $\pi_k$ (\resp, $\bm{\mu}_k$ and $\bm{\sigma}_k^2$) stands for the prior (\resp, mean and diagonal covariance) of the $k$-th component in GMM.
We define the probability that the $i$-th region proposal belongs to the $j$-th Gaussian model as
\begin{eqnarray}
\gamma_i(j) = \frac{\pi_j\mathcal{N}(\x_i;\bm{\mu}_j, \textnormal{diag}(\bm{\sigma}_j^2))}{\sum_{k=1}^K \pi_k\mathcal{N}(\x_i;\bm{\mu}_k, \textnormal{diag}(\bm{\sigma}_k^2))}.
\end{eqnarray}

With $\bm{\gamma}_i=[\gamma_i(1),\ldots,\gamma_i(K)]$, we calculate the average of $\bm{\gamma}_i$'s of all the region proposals from the images belonging to the $c$-th category as the encoding vector of the $c$-th category, which is denoted as $\bar{\bm{\gamma}}_c$. 
Thus, each category can be represented by a $K$-dim visual encoding vector. However, there are two remaining issues to be addressed.

The first issue is that the region proposals from one category may be very noisy due to inaccurate image labels. In each visual encoding vector $\bar{\bm{\gamma}}_c$, the small entries are more likely to be induced by outliers. Thus, we set the $10$ smallest entries in each $\bar{\bm{\gamma}}_c$ as $0$ to suppress the negative impact of label noise.

The second issue is that some components in GMM (\ie, visual words in the codebook) are category-invariant and commonly shared by all categories (\eg, visual words that fall in the background). In order to obtain more discriminative visual encoding vectors, we tend to learn an orthonormal transformation matrix $\W\in\mathcal{R}^{K\times K}$ to maximize the category separation, \ie, $\frac{1}{2}\sum_{c\neq \tilde{c}} \|\W\bar{\bm{\gamma}}_c-\W\bar{\bm{\gamma}}_{\tilde{c}}\|^2$.
By defining $\R\in\mathcal{R}^{K\times C}$ with the $c$-th column being $\bar{\bm{\gamma}}_c$ and $\H\in\mathcal{R}^{C\times C}$ is a Laplacian matrix with the diagonal elements as $C-1$ and the remaining elements as $-1$, we arrive at the following optimization problem:
\begin{eqnarray} \label{eqn:max_separation2}
\max_{\W:\,\W\W^T = \I} \textnormal{trace}(\W\R\H\R^T\W^T),
\end{eqnarray}
which can be solved by combining the $K$ leading eigen-vectors of $\R\H\R^T$ as $\W$. However, fast decay of eigen-values may result in the low quality of $\W$. Especially when $C<K$, $K-C$ eigen-values of $\R\H\R^T$ are $0$. Hence, we solve the following relaxed problem of (\ref{eqn:max_separation2}) instead:
\begin{eqnarray} \label{eqn:max_separation_relax}
\max_{\W}\,\,\textnormal{trace}(\W\R\H\R^T\W^T)\!-\!\beta\|\W\W^T-\I\|_F,
\end{eqnarray}
in which $\beta$ is a trade-off parameter and empirically set as $100$ in our experiments. Inspired by \cite{yu2013designing}, after initializing $\W$ as an empty matrix, we incrementally learn an additional row of $\W$. Particularly, in the $k$-th iteration, given $\W^k$, we aim to learn the $(k+1)$-th row $\w$. By assuming $\w\w^T=1$, the subproblem in (\ref{eqn:max_separation_relax}) \wrt $\w$ can be written as 
\begin{eqnarray}
\max_{\w:\,\w\w^T=1} \w(\R\H\R^T-2\beta \W^T\W)\w^T,
\end{eqnarray}
which can be easily solved by seeking the leading eigen-vector of $(\R\H\R^T-2\beta \W^T\W)$ with the largest eigen-value. After obtaining $\w$, we update $\W^{k+1}=[\W^k; \w]$. To obtain $\tilde{K}$ leading rows of $\W$, we repeat the above procedure for $\tilde{K}$ times. To this end, we obtain our free and visually grounded visual encoding for each category, which is robust \wrt noisy image labels and rich in discriminative information.

In practice, among the three types of semantic representations mentioned above, we can concatenate arbitrary number of available types as the hybrid semantic representation for each category. More experimental details will be described in Section~\ref{sec:exp}.

\section{Experiments} \label{sec:exp}
In this section, we evaluate our WSCI method for image classification on three benchmark datasets. Extensive experimental results together with in-depth qualitative and quantitative analyses demonstrate the effectiveness of our proposed method.

\noindent\textbf{Datasets: }Since attribute vector is part of our hybrid semantic information, we conduct experiments on three popular benchmark datasets: AwA2, CUB, and SUN Attribute, which are associated with attribute vector for each category and widely used for zero-shot learning~\cite{akata2015evaluation, frome2013devise, socher2013zero, yu2013designing}. For each benchmark dataset, we use the whole dataset as test set and collect the training set by crawling $500$ web images per category from Google image website.
\begin{itemize}
\item AwA2~\cite{xian2017zero}:
The Animals with Attributes 2 (AwA2) dataset is a drop-in replacement of original Animals with Attributes (AwA) dataset, with more images released for each category. Specifically, AwA2 consists of in total $37322$ images distributed in $50$ animal categories. The AwA2 also provides category-attribute matrix, which contains an $85$-dim attribute vector (\eg, color, stripe, furry, size, and habitat) for each category.

\item CUB~\cite{WahCUB_200_2011}: The Caltech-UCSD Bird (CUB) dataset consists of in total $11,788$ images from $200$ bird species. One $312$-dim binary human specified attribute vector (\eg, bill, wing, tail, eye, and belly) is provided for each image, so we average the attribute vectors of the images within each category and utilize the averaged attribute vector for that category.

\item SUN Attribute~\cite{xiao2010sun}: Scene UNderstanding (SUN) attribute dataset has $717$ scene categories with $20$ images in each category. 
Similar to CUB, a $102$-dim attribute vector (\eg, function, material, appearance, and property) is provided for each image, and thus we calculate the averaged attribute vector for each category.

\item Google image dataset: We collect the web training set by ourselves. Specifically, for each benchmark test set (\ie, AwA2, CUB, and SUN), we crawl the top ranked $500$ images from Google image website for each category by using the category name as query. Besides, we perform PCA-based near-duplicate removal~\cite{zhou2016places} to guarantee that the web training set has no overlap with the test set. 
\end{itemize}

\noindent\textbf{Category-level Semantic Representation: }
As discussed in Section~\ref{sec:csi}, we utilize three types of category-level semantic information: attribute, word vector, and visual encoding. Next, we provide the experimental details of extracting these three types of information: 
\begin{itemize}
\item For attribute, we use the $85$-dim (\resp, $312$-dim and $102$-dim) attribute vector associated with the AwA2 (\resp, CUB and SUN) dataset as mentioned above.

\item For word vector, we train GloVe~\cite{pennington2014glove} language model based on the latest Wikipedia corpus, with the dimension of word vector set as $500$. Then, we can obtain the word vector corresponding to each category name. Note that some category names contain more than one word. In this case, we average the word vectors corresponding to all the words appearing in the category name as the final word vector for that category.

\item For visual encoding, $K$ is set as $256$ following~\cite{sanchez2013image} and  $\tilde{K}$ is set as $128$ to reduce the dimension of visual encoding vector by half. Thus, the dimension of final visual encoding is $128$. 
\end{itemize}

Three types of semantic representations are concatenated as hybrid semantic representation, yielding a $713$-dim (\resp, $940$-dim and $730$-dim) vector for each category on the AwA2 (\resp, CUB and SUN) dataset.

\noindent\textbf{Network Architecture: }As illustrated in Figure~\ref{fig:flowchart}, our system consists of a CNN model, a VAE model, and a softmax classification model. For the CNN model, we use Inception-V3~\cite{szegedy2016rethinking}, which outputs $2048$-dim visual feature. For the VAE model, we implement both encoder and decoder as multiple layer perceptron (MLP) with one hidden layer. The dimension of the hidden layer in MLP is set as $1500$, which is approximately $\frac{d+m}{2}$ with $d$ being the dimension of Inception-V3 output (\ie, $2048$) and $m$ being the dimension of category-level semantic representation. In the training process, we use Adam optimizer for optimization with batch size fixed as $64$ and exponentially decaying learning rate initialized as $0.001$. The entire system is implemented by using TensorFlow.

\noindent\textbf{Parameter: }The objective function in (\ref{eqn:final_obj}) has one hyper-parameter $\tilde{\lambda}$, which is empirically set as $10^{-4}$ on all three datasets. In fact, we observe that our method is relatively robust with $\tilde{\lambda}$ set within certain range (\eg, [$10^{-6}$,$10^{-4}$]) in our experiments.

\setlength{\textfloatsep}{5pt}
\begin{table}[t]
\caption{Accuracies (\%) of different methods on three datasets. The best results are highlighted in boldface.}
\setlength{\tabcolsep}{5pt}
\label{tab:exp_results}
\centering
\begin{tabular}{|c|c|c|c|c|}
\hline
Dataset & AwA2 & CUB & SUN & Avg \\
\hline
CNN  & 84.02 & 72.24 & 35.91 & 64.06 \\
\hline
bootstrap~\cite{reed2014training} & 85.71 & 73.63 & 37.36 & 65.57\\ 
Chen and Gupta~\cite{chen2015webly} & 85.53 & 74.92 & 38.33 & 66.26\\ 
Sukhbaatar~\etal~\cite{sukhbaatar2014training}  & 86.11 & 73.51 & 38.61 & 66.08\\
Xiao~\etal~\cite{xiao2015learning} & 86.41 & 75.02 & 40.56 & 67.33\\
RGT+AT+R~\cite{zhuang2016attend} & 86.52 & 73.75 & 38.03 & 66.10\\
\hline
WSCI\_sim1  & 86.15 & 74.17 & 37.91 & 66.08\\
WSCI\_sim2  & 88.88 & 76.28 & 41.23 & 68.80\\
\hline
WSCI (w/o attr)  & 89.56 & 76.61 & 41.03 & 69.07\\
WSCI (w/o ve)  & 90.52 & 76.86 & 41.97 & 69.78\\
WSCI  & \textbf{91.14} & \textbf{77.34} & \textbf{42.26} & \textbf{70.25}\\
\hline
\end{tabular}
\end{table}

\setlength{\textfloatsep}{5pt}
\begin{figure*}[t]
        \centering
        \begin{subfigure}[b]{1.0\textwidth}     
        \centering   
                \includegraphics[width=0.49\textwidth]{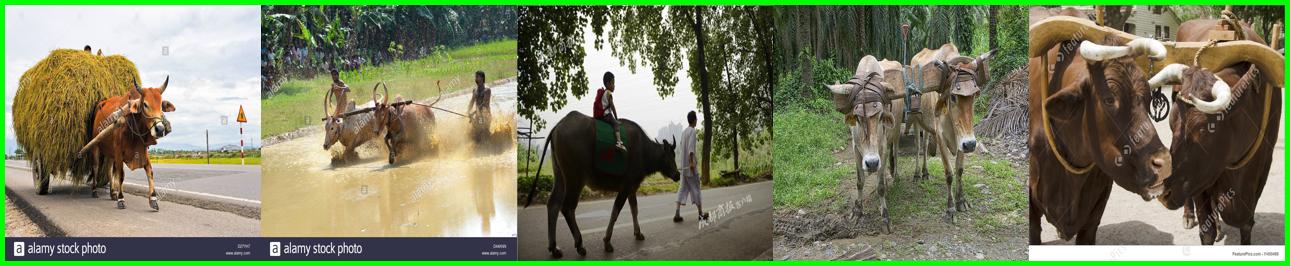}
                \includegraphics[width=0.49\textwidth]{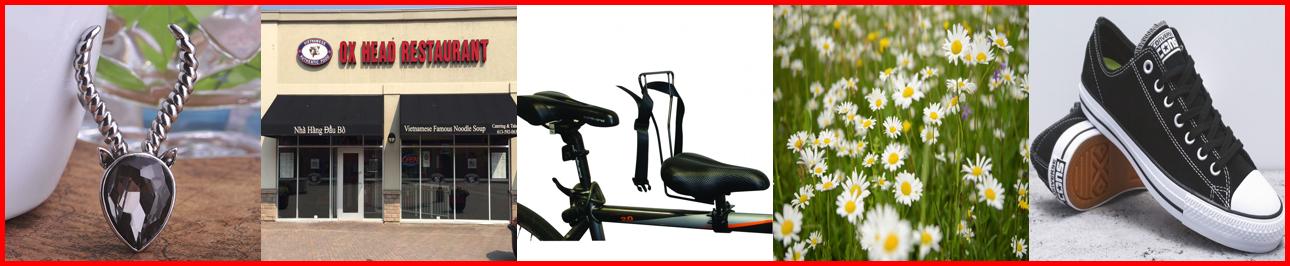}         
        \vspace{3pt}     
        \end{subfigure}
        \begin{subfigure}[b]{1.0\textwidth}     
        \centering   
                \includegraphics[width=0.49\textwidth]{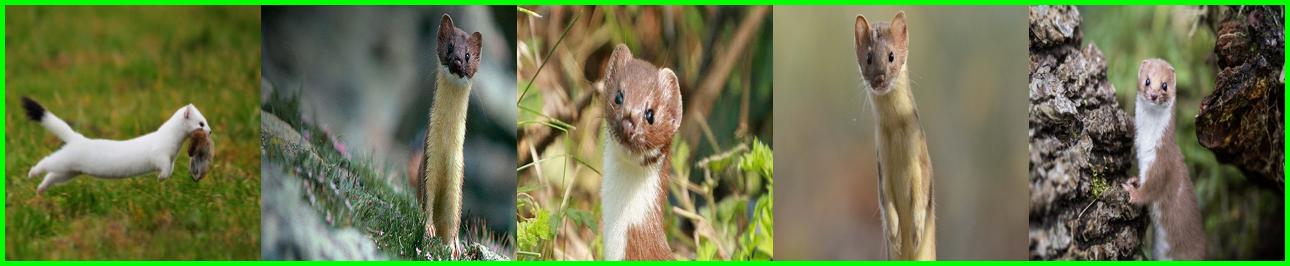}
                \includegraphics[width=0.49\textwidth]{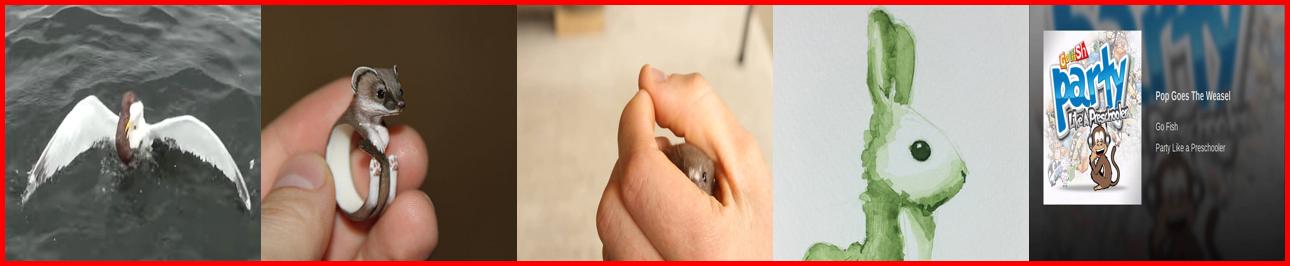}              
        \end{subfigure}
        \caption{The first (\resp, second) row contains the web training images from the category ``ox" (\resp, ``weasel").
        The green (\resp, red) boxes on the left (\resp, right) column group the top $5$ images with highest (\resp, lowest) $\tilde{p}_{\bm{\theta}_1}(\x^i|\z^i)$, which indicates the non-outliers (\resp, outliers) identified by our WSCI method.} 
        \label{fig:outliers}
\end{figure*}

\noindent\textbf{Baselines: }For baselines, we compare with three sets of baselines: basic CNN, webly supervised learning (WSL) methods, and simplified versions of our WSCI method. 

\noindent 1) For basic CNN, we train Inception-V3 without accounting for label noise and refer to this basic baseline as CNN in Table~\ref{tab:exp_results}. 

\noindent 2) For WSL baselines, we compare with the following recent deep learning methods: Chen and Gupta~\cite{chen2015webly}, Sukhbaatar~\etal~\cite{sukhbaatar2014training}, Xiao~\etal~\cite{xiao2015learning}, bootstrap~\cite{reed2014training}, and RGT+AT+R~\cite{zhuang2016attend}. For Chen and Gupta~\cite{chen2015webly} and Xiao~\etal~\cite{xiao2015learning} which require confusion matrix estimated based on clean data, due to lack of clean data, we estimate the confusion matrix by calculating the similarities among categories based on semantic representations. In this sense, category-level information is also used in~\cite{chen2015webly,xiao2015learning}. For Xiao~\etal~\cite{xiao2015learning} and RGT+AT+R~\cite{zhuang2016attend}, partial clean data are need when training the network, since clean data are not available in our scenario, we only train their models by using noisy web data for fair comparison. In particular, three web images constitute each training bag without using annotated clean image. For all the baselines involving deep network, we use Inception-V3 as the basic CNN structure, to eliminate the possibility that the performance improvement is attributed to advanced CNN structure instead of our design.

\noindent 3) For simplified versions of our WSCI method, we compare with the following special cases for ablation study:
\begin{itemize}
\item We first split the flowchart in Figure~\ref{fig:flowchart} into two separate flows with joint loss at the end, in which the top flow is classification network based on $\x$ and the bottom flow is a plain VAE detached from the classifier.  We use the same loss function as in (\ref{eqn:final_obj}) and refer to this simplified version as WSCI\_sim1 in Table~\ref{tab:exp_results}. Note that the plain VAE detached from the classifier does not utilize category-level semantic representations, so category-level  information is not included in WSCI\_sim1.
\item Based on WSCI\_sim1, we replace VAE with semantic VAE by attaching a new classifier on the hidden layer, which can utilize category-level information $\A$. Note that the newly attached classifier has an additional standard classification loss, which is independent on the classification network in the top flow. This simplified version is referred to as WSCI\_sim2, in which the classification network in the top flow and the semantic VAE in the bottom flow cannot jointly utilize the category-level semantic information.
\item To demonstrate the effectiveness of our proposed visual encoding, we exclude visual encoding from hybrid semantic representation, which is referred to as WSCI (w/o ve). Note that WSCI (w/o ve) is nearly the same as full-fledged WSCI except missing the visual encoding.
\item In some real-world applications, manually annotated attribute vectors are not always available. To clear this concern, similar to WSCI (w/o ve), we exclude attribute vector from hybrid semantic representation, which is dubbed WSCI (w/o attr).

\end{itemize}

\noindent\textbf{Experimental Results: }
The experimental results are summarized in Table~\ref{tab:exp_results}, in which the results on the SUN dataset are much worse than those on the other two datasets. This is because the SUN dataset has far more categories and the web images of scene categories are more noisy than those of animal categories. Besides, we have the following observations:
\begin{itemize}
\item The baselines~\cite{reed2014training,chen2015webly,sukhbaatar2014training,xiao2015learning,zhuang2016attend} outperform basic CNN, because they account for the label noise of web data via different approaches. 

\item WSCI\_sim2 is better than WSCI\_sim1, which indicates the benefit of utilizing category-level information. 

\item WSCI achieves better results than WSCI\_sim2, which validates the advantage for classification network and VAE to jointly leverage category-level information.

\item WSCI is slightly better than WSCI (w/o ve), which shows that it is useful to include our proposed visual encoding vector as part of hybrid semantic representation. 

\item In the absence of manually annotated attribute vector, our method WSCI (w/o attr) is still superior compared with all the baselines, which demonstrates the effectiveness of our method even only using free category-level semantic information.

\item Our WSCI method outperforms all the baselines and achieves the best results on all three datasets, which shows the superiority of our VAE based method with VAE and classification network jointly utilizing category-level supervision.
\end{itemize}

\noindent\textbf{Qualitative Analysis: }
Besides quantitative experimental results, we also would like to verify the ability of our semantic VAE for outlier detection qualitatively. 
Recall that the reconstruction probability density $\tilde{p}_{\bm{\theta}_1}(\x^i|\z^i)$ in (\ref{eqn:final_obj}) is used as the indicator of $\x^i$ being a non-outlier, in which higher (\resp, lower) $\tilde{p}_{\bm{\theta}_1}(\x^i|\z^i)$ corresponds to non-outliers (\resp, outliers). By taking the categories ``ox" and ``weasel" from the dataset AwA2 as examples (see the top and bottom row in Figure~\ref{fig:outliers}), we show the top $5$ web training images from each category with lowest (\resp, highest) $\tilde{p}_{\bm{\theta}_1}(\x^i|\z^i)$ obtained from our WSCI method in the red (\resp, green) boxes. From Figure~\ref{fig:outliers}, we observe that our WSCI method succeeds in identifying the top non-outliers and outliers, which contributes to learning a more robust classifier. We have similar observations on the other categories and the other datasets.

\noindent\textbf{Different Noise Levels: }To explore whether our WSCI method can consistently outperform baselines \wrt different noise levels, we conduct additional experiments by using different training sets with different ratios of noisy images.
In previous works on learning from noisy training data, noise is generally manually injected into a clean dataset by flipping the labels for precise control of noise ratio. However, our web training data are not associated with accurate labels, and hence we cannot control the noise ratio precisely by flipping the labels of training data. Therefore, we adopt an alternative approach, \ie, using the order returned by the Google searching engine to simulate different noise levels, since it is usually assumed that the top retrieved example are of high quality and more likely to non-outliers. Particularly, we use $s(i)$ to indicate the setting, in which the training images start from index $i$ and end with index $i + 499$ for each category based on the searching engine (recall that we use $500$ web training images for each category). Note that $s(1)$ is the setting reported in Table~\ref{tab:exp_results}. By taking the SUN dataset as an example, the accuracies (\%) of the strongest baseline Xiao~\etal~\cite{xiao2015learning} and our WSCI method under the settings $s(1)$, $s(201)$, $s(401)$ are reported in Table~\ref{tab:exp_noise_levels}, from which we can observe that
the performances of both methods decrease as the noise ratio arises. However,
our WSCI method outperforms Xiao~\etal~\cite{xiao2015learning} under all three settings, which demonstrates the consistent superiority of our WSCI method compared with baselines \wrt different noise levels. We have similar observations on the other datasets.

\setlength{\textfloatsep}{5pt}
\begin{table}[t]
\caption{Accuracies (\%) of different methods under different settings with different noise levels on the SUN dataset. The best results are highlighted in boldface.}
\setlength{\tabcolsep}{5pt}
\label{tab:exp_noise_levels}
\centering
\begin{tabular}{|c|c|c|c|}
\hline
Setting & s(1) & s(201) & s(401) \\
\hline
Xiao~\etal~\cite{xiao2015learning} & 40.56 & 39.85 & 39.00  \\
\hline
WSCI & \textbf{42.26} & \textbf{41.79} & \textbf{41.47} \\ 
\hline
\end{tabular}
\end{table}

\section{Conclusion} \label{sec:conclusion}
In this work, we have explored addressing the label noise by utilizing category-level supervision when learning from web data. 
In particular, we have proposed our WSCI method, in which VAE and classification network can jointly leverage category-level information to account for label noise. Comprehensive experiments on three benchmark datasets have demonstrated the effectiveness of our proposed method.

\bibliographystyle{IEEEtran}
\bibliography{VAE_bib}

\end{document}